\documentclass[nohyperref]{article}

\usepackage{microtype}
\usepackage{graphicx}
\usepackage{booktabs} 
\usepackage{url}            
\usepackage{amsfonts}       
\usepackage{placeins}
\usepackage{nicefrac}       
\usepackage{xcolor}         
\usepackage{float}
\usepackage{graphicx}
\usepackage{subcaption}
\usepackage{vcell}
\usepackage{multirow}
\usepackage{array}
\usepackage{listings}
\usepackage{tabularx}
\usepackage{makecell}
\usepackage[utf8]{inputenc} 
\usepackage[T1]{fontenc}    
\usepackage{rotating}

\usepackage{hyperref}

\usepackage[accepted]{icml2022}

\usepackage{amsmath}
\usepackage{amssymb}
\usepackage{mathtools}
\usepackage{amsthm}

\usepackage[capitalize,noabbrev]{cleveref}

\theoremstyle{plain}

\theoremstyle{definition}

\theoremstyle{remark}

\usepackage[textsize=tiny]{todonotes}

\icmltitlerunning{BERT-JEPA}

\begin{document}

\twocolumn[
\icmltitle{BERT-JEPA: Reorganizing CLS Embeddings for Language-Invariant Semantics}



\icmlsetsymbol{equal}{*}

\begin{icmlauthorlist}
\icmlauthor{Taj Gillin}{equal,brown}
\icmlauthor{Adam Lalani}{equal,brown}
\icmlauthor{Kenneth Zhang}{equal,brown}
\icmlauthor{Marcel Mateos Salles}{equal,brown}
\end{icmlauthorlist}

\icmlaffiliation{brown}{Department of Computer Science, Brown University, Providence, RI, USA}

\icmlcorrespondingauthor{Marcel Mateos Salles}{marcel\_mateos\_salles@brown.edu}

\icmlkeywords{Machine Learning, ICML}

\vskip 0.3in
]



\printAffiliationsAndNotice{\icmlEqualContribution} 

\begin{abstract}
    Joint Embedding Predictive Architectures (JEPA) are a novel self supervised training technique that have shown recent promise across domains. We introduce BERT-JEPA (BEPA), a training paradigm that adds a JEPA training objective to BERT-style models, working to combat a collapsed [CLS] embedding space and turning it into a language-agnostic 
    space. 
    This new structure leads to increased performance across multilingual benchmarks.
    
\end{abstract}

\section{Introduction}
\label{introduction}

Language models like BERT \citep{devlin2019bertpretrainingdeepbidirectional} and its successors are used to generate rich embeddings for NLP tasks. For many downstream tasks, BERT embeddings can easily be taken out of the box and leveraged to achieve competitive performance. However, these embeddings fail to capture a true representation behind language. Its [CLS] token, one of the most commonly applied to downstream tasks, has been found to not adequately capture the meaning of sentences for sentence similarity tasks \citep{reimers2019sentencebertsentenceembeddingsusing}. Language is a concrete manifestation of thoughts that preserve meaning with varying syntax and structure. This is reflected by a basic translation between Spanish and English: "El gato es rojo" and "The cat is red". They contain and convey the same intention and thoughts through separate linguistic systems. Therefore, we ask the question: Can BERT be taught to represent information independent of the linguistic system, only representing the true meaning behind the sentences and information? We turn to the emerging literature on Joint-Embedding Predictive Architectures (JEPA), which focuses on predicting latent representations between two different samples \citep{lecue2022path}. Recently, these architectures have shown promise in vision \citep{assran2023selfsupervisedlearningimagesjointembedding, assran2025vjepa2selfsupervisedvideo, bardes2024revisitingfeaturepredictionlearning} and language tasks, specifically code-text tasks \citep{huang2025llmjepalargelanguagemodels}. This naturally extends to general language, where a model can learn to predict the embedding of a sample from another with the same meaning, encouraging a shared embedding space, which we have termed the "thought space" of the model. This rich embedding space corresponds to the abstraction above language that humans have, our thoughts. We reveal our new training paradigm, BERT-JEPA (BEPA), which aims to build a rich contextual understanding by leveraging JEPA. We get the following results:
\begin{enumerate}
    \item \textbf{BEPA finetuning drastically reorganizes the [CLS] embedding space to a semantic-first structure.}
    \item \textbf{BEPA finetuning shifts the PCA representation from low-rank to fuller-rank.}
    \item \textbf{[CLS] embedding reorganization comes at little to no loss in English performance.}
    \item \textbf{BEPA finetuning increases performance in multilingual tasks.}
\end{enumerate}

\section{Background}
\label{background}

The field of NLP has changed significantly in the last six years. The release of BERT \citep{devlin2019bertpretrainingdeepbidirectional} revolutionized the field of language understanding by leveraging the emerging transformer architecture \citep{vaswani2023attentionneed} to build a deep network that could generate rich embeddings for downstream tasks. BERT released new pretraining tasks, the most important being its masked language modeling (MLM) task which allows for BERT's bidirectional attention to function properly.

Follow-up models such as \textit{RoBERTa} \citep{liu2019robertarobustlyoptimizedbert} and \textit{XLM-RoBERTa} \citep{conneau2018xnlievaluatingcrosslingualsentence} were improvements to BERT, that utilized longer training and removed unnecessary training tasks such as next sentence prediction (NSP) and translation language modeling (TLM). By focusing purely on building a strong latent representation, these models were able to significantly increase performance across benchmarks when compared to base BERT. We will use \textit{RoBERTa} and \textit{XLM-RoBERTa} as baselines and comparisons in our experiments. 

Despite this progress, language models have failed to become language invariant \citep{ribeiro-etal-2018-semantically}, as small semantic preserving changes to examples can cause a model to change its behavior and predictions. This means that language models tend to still encode language-specific features.

Joint-Embedding Predictive Architectures (JEPA) have changed the way models are trained. Rather than aligning a model in the input space, \citet{lecue2022path} proposed an architecture that would align a model in embedding space, creating a more robust representation and understanding. In practice, these architectures are scalable and have performed extremely well in vision tasks, competing with and outperforming existing methods across downstream tasks \citep{assran2023selfsupervisedlearningimagesjointembedding, assran2025vjepa2selfsupervisedvideo, bardes2024revisitingfeaturepredictionlearning}. The embeddings generated from these models are context-rich and made up of the most impactful features \citep{littwin2024jepaavoidsnoisyfeatures}. Recently, a JEPA variation was introduced for language tasks \citep{huang2025llmjepalargelanguagemodels}, achieving impressive performance and lending itself well to NL-Regex and NL-SQL cases. We were inspired by this paper to apply JEPA to build richer embedding spaces in BERT and its variants. 

To assess the quality of our new embeddings we plan to use a variety of benchmarks and techniques. Commonly, BERT and its variants are benchmarked on SST-2 \citep{socher-etal-2013-recursive}. Furthermore, the STS Crosslingual datasets \citep{cer-etal-2017-semeval, enevoldsen2025mmtebmassivemultilingualtext, muennighoff2022mteb} can be used to understand if the semantic representations are aligned across languages and BEPA is able to abstract away language. In addition to down-stream evaluations, we can utilize diagnostic methods such as t-SNE visualization \citep{cai2022theoreticalfoundationstsnevisualizing} and isotropy \citep{mickus2024isotropyclustersclassifiers} to study properties of our embeddings. We plan to use all these techniques to evaluate BEPA's performance.

\section{Method: A Novel JEPA-style BERT Finetuning Framework}
\label{methods}

In this section, we cover the training regimen used to train BERT-style models utilizing BERT-JEPA (BEPA). For complete framework details, see \cref{full bepa framework}. Training details and hyperparameters used can be found in \cref{resources} and the corresponding repository and details is in \cref{repo}. We begin by sharing the training framework and then dive deeper into the losses used.

\subsection{BEPA Framework}

Our novel BERT training method transforms the current embedding space for BERT models to make them invariant to linguistic systems and fix known CLS-token collapse \citep{reimers2019sentencebertsentenceembeddingsusing}. We begin by taking a base pretrained \textit{RoBERTa} model (\textit{xlm-roberta-base}) \citep{DBLP:journals/corr/abs-1911-02116} but any BERT-style model works. We embed two sentences side by side, packing them into a single transformer input. It is tokenized with special tokens added as seen in \cref{tokenized input}. Individually, we tested two different methods for this setup: \textit{monolingual} and \textit{bilingual}. In which \textit{monolingual} only packaged samples in the same language side by side and \textit{bilingual} allowed samples of different languages to be packaged together. The main paper focuses on results for the \textit{bilingual} packaging strategy, results for \textit{monolingual} and other ablations can be found in \cref{more results}.
This setup then undergoes BERT's classic MLM training \citep{devlin2019bertpretrainingdeepbidirectional}. We do this to preserve  strong linguistic capabilities of BERT-style transformers.

We then move on to the novel part of the BEPA framework. Because BERT-style models treat the final-layer [CLS] token as the latent representation of the sequences, we take the [CLS] tokens and compute an alignment loss that encourages invariance across languages. We conducted two separate forward passes with each individual sentence masked out as seen in \cref{BEPA masking}. The latent representation of each sentence's [CLS] token is then taken and aligned through $L_{Alignment}$, more information can be found in \cref{BEPA loss}. We leveraged an identity predictor before aligning [CLS] tokens, but future work should test other type and styles of predictors. A diagram of this step can be found in \cref{bepa architecture}.

\subsection{BEPA Loss}
\label{BEPA loss}

The loss we chose to use can not be a basic loss, as it has to account for the dual training objectives that are used for BEPA. BERT-style MLM and CLS-token JEPA alignment result in a new training objective of:
$$L_{BEPA} = L_{MLM} + \lambda L_{Alignment}$$
in which $\lambda$ is a hyperparameter determining how much importance to put on $L_{Alignment}$. We utilize Cross Entropy, similar to a typical BERT-style model as $L_{MLM}$ and for $L_{Alignment}$, we found InfoNCE \citep{oord2019representationlearningcontrastivepredictive} to be the most effective. Future work should test the impact of different objectives such as SigREG \citep{balestriero2025lejepaprovablescalableselfsupervised}. The framework is fully extensible and current losses can be swapped out with whatever loss the user requires, we found Mean Squared Error (MSE) and Cosine Similarity to not be as effective for the alignment task, as they do not fight the [CLS] collapse.

\section{BEPA Training Semantically Reorganizes the CLS Embedding Space}
\label{fights collapse}

In this section, we study the resulting [CLS] space post BEPA finetuning. Before finetuning, BERT-style models, displayed a collapsed embedding space. This can be seen in \cref{collapsed}, especially in \cref{collapsed-cls-full-roberta,collapsed-cls-full-xlm}, where any pairs, semantically similar or not, displayed high cosine similarities. This trend was seen across both regular \textit{RoBERTa} and \textit{XLM-RoBERTa}. 

However, after BEPA finetuning, a different story appears. Our models began to display high alignment for semantically similar pairs (both within and across languages) and lower degrees of alignment for semantically different pairs (both within and across languages). To see this trend, refer to \cref{fig:pos-neg} and \cref{new embedding space} in the appendix.

\begin{figure}[H]
  \centering
  \includegraphics[width=0.4\textwidth]{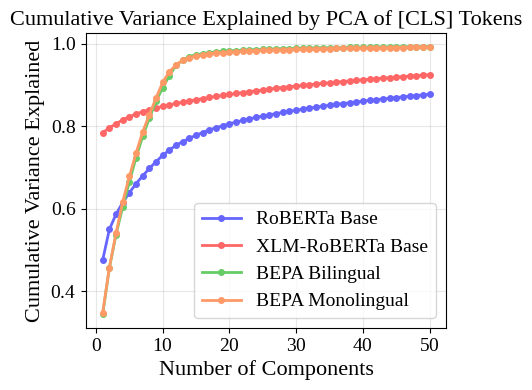}\caption{ PCA plot showing the variance per component of the [CLS] embedding space. \textbf{BEPA finetuning increases (↑) the number of high variance components.} Meanwhile, standard \textit{RoBERTa} and \textit{XLM-RoBERTa} display the majority of variance, 47\% and 78\%  respectively, being displayed by a \textbf{single} component. Compared to 34\% in the first component for both BEPA models.}
  \label{PCA plot}
\end{figure}

To further investigate the reorganized [CLS] embedding, we ran PCA to explain the variance per component of the space. We find that the \textit{monolingual} and \textit{bilingual} models distribute variance over a larger number of principal components, indicating that more embedding dimensions are informative and thus their [CLS] representations are less redundant. This finding can be seen in \cref{PCA plot}.

\begin{figure*}
  \centering
  \hspace*{-0.06\textwidth}
  \includegraphics[width=1.1\textwidth]{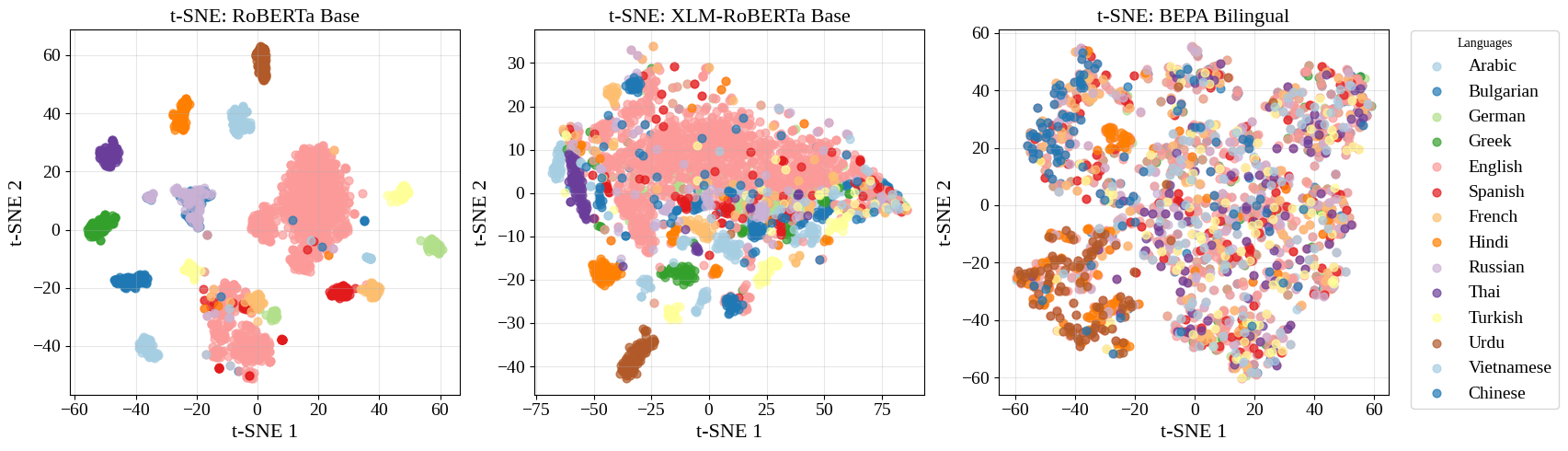}
  \caption{\textit{(Left)}: t-SNE plot for \textit{RoBERTa}, samples are tightly distributed by language, with English distinctly unique in its large span.  \textit{(Middle)}: t-SNE plot for \textit{XLM-RoBERTa}, some languages are isolated, similar to \textit{RoBERTa}, though less tightly packed. Many languages, especially English, span a much larger space and share a space with multiple languages.  \textit{(Right)}: t-SNE plot for \textit{BEPA Bilingual}, languages are evenly distributed in overlapping. \textbf{BEPA creates a shared and aligned space between languages}. Results for our finetuned \textit{BEPA Monolingual} model can be found in \cref{tsne_plot_mono}.}
  \label{fig:tsne_plot}
\end{figure*}

Furthermore, t-SNE visualizations of the [CLS] embeddings show pronounced language-specific clusters for \textit{RoBERTa} and \textit{XLM-RoBERTa}, whereas our \textit{monolingual} and \textit{bilingual} models exhibit substantially more overlap across languages, indicating a more language-invariant embedding space. These clusters can be found in \cref{fig:tsne_plot}.


\begin{table*}[t]
\centering
\caption{Average accuracy (decimal) across 5 seeds on XNLI Language Transfer benchmark, displayed by language and overall. Results are gathered by finetuning on English and then doing zero-shot evaluation on each language. \textbf{\textit{BEPA Bilingual} consistently outperforms (↑) \textit{XLM-RoBERTa} across all languages.} For additional results and ablations refer to \cref{tab:xnli_results_app,more results}.}
\label{tab:xnli_results}
\resizebox{\textwidth}{!}{%
\begin{tabular}{|l|ccccccccccccccc|c|}
\toprule
\textbf{Model} & \textbf{ar} & \textbf{bg} & \textbf{de} & \textbf{el} & \textbf{en} & \textbf{es} & \textbf{fr} & \textbf{hi} & \textbf{ru} & \textbf{sw} & \textbf{th} & \textbf{tr} & \textbf{ur} & \textbf{vi} & \textbf{zh} & \textbf{Avg} \\
\midrule
XLM-RoBERTa & 0.705 & 0.753 & 0.753 & 0.744 & 0.831 & 0.783 & 0.767 & 0.687 & 0.740 & 0.610 & 0.714 & 0.706 & 0.656 & 0.735 & 0.741 & 0.728 \\
BEPA Bilingual & \textbf{0.717} & \textbf{0.765} & \textbf{0.771} & \textbf{0.763} & \textbf{0.842} & \textbf{0.794} & \textbf{0.780} & \textbf{0.693} & \textbf{0.750} & \textbf{0.673} & \textbf{0.726} & \textbf{0.725} & \textbf{0.657} & \textbf{0.749} & \textbf{0.751} & \textbf{0.744} \\
BEPA Monolingual & 0.707 & 0.755 & 0.755 & 0.745 & 0.829 & 0.782 & 0.771 & 0.681 & 0.738 & 0.662 & 0.714 & 0.718 & 0.654 & 0.735 & 0.735 & 0.732 \\
\bottomrule
\end{tabular}%
}
\end{table*}

\section{Semantically Organized Token Spaces
Improve Cross-Lingual Transfer} 

In this section, we cover the performance of our BEPA finetuned models across benchmarks. We test on GLUE, XNLI, and MLQA. The complete results of these benchmarks can be found in \cref{more results}.

We ran the XNLI benchmark to test whether BEPA finetuning would help a BERT-style model to understand sentence-level semantics and apply it across languages. If these BEPA finetuned models are reasoning through a semantically aligned "thought space", then they should be able to more accurately perform across languages. To run the XNLI benchmark, we finetuned for two epochs on the English language and then conducted zero shot evaluation of all the other languages. Results for our models can be found in \cref{tab:xnli_results,tab:xnli_results_app}. We find that BEPA finetuning (specifically with the \textit{Bilingual} packing set up) leads to consistent increased performance across all languages, highlighting a better language-invariant and semantically aligned [CLS] embedding space.

\begin{figure}[H]
  \centering
  \includegraphics[width=0.505\textwidth]{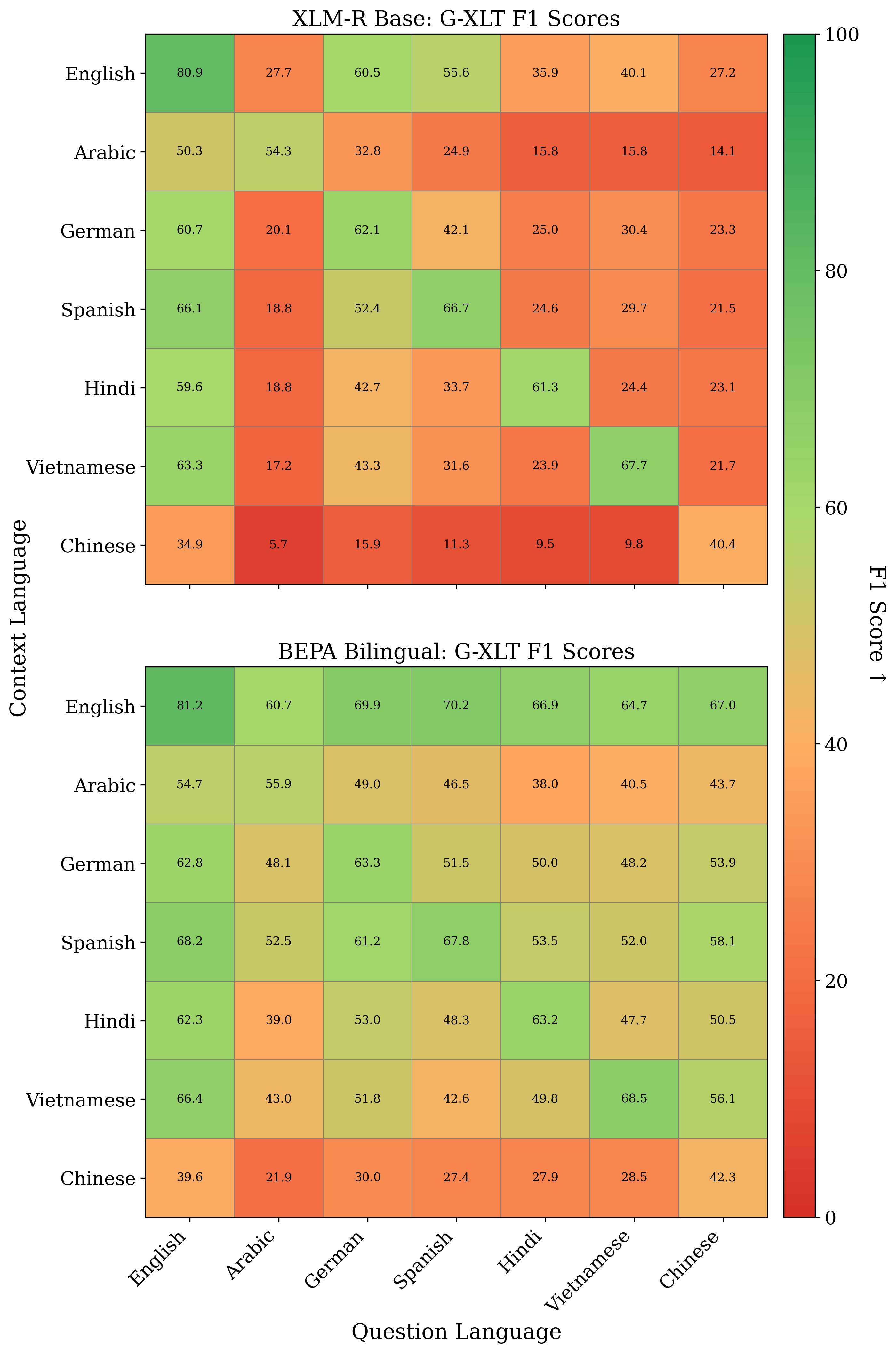}
 \caption{Performance G-XLT matrix on MLQA benchmark. F1 scores across 49 language pairs (7 context languages × 7 question languages). Diagonal entries represent monolingual performance, while off-diagonal entries show cross-lingual transfer. \textit{(Top)}:  \textit{XLM-RoBERTa} Base shows strong performance on monolingual tasks but struggles significantly with cross-lingual scenarios, particularly those not involving English. \textit{(Bottom)}: \textit{BEPA Bilingual} demonstrate improved cross-lingual transfer, with more uniform performance across language pairs, while maintaining strong monolingual performance. This indicates \textbf{BEPA's learns token-level semantic alignment in the representation space}.}
  \label{MLQA}
\end{figure}

We ran the MLQA (Multilingual Question Answering) benchmark to test whether our BEPA finetuning would help a BERT-style model understand and extract information at the token-level across languages. Unlike XNLI which evaluates sentence-level classification, MLQA requires extractive question answering where the model must identify specific answer spans within a context paragraph. If BEPA finetuned models achieve better semantic alignment in their representation space, they should be able to transfer token-level reasoning capabilities across languages more effectively. To run the MLQA benchmark, we finetuned for two epochs on English SQuAD v1.1 \citep{rajpurkar2016squad} and then conducted zero-shot evaluation on all 49 language pairs ($7 \times 7$ G-XLT matrix). We report F1 scores for monolingual (same context and question language) and cross-lingual (different context and question languages) configurations. Results are shown in \cref{MLQA}. While we see good and similar performance on same monolingual configurations for both the \textit{XLM-RoBERTa} baseline and our \textit{BEPA Bilingual} model, our \textit{BEPA Bilingual} model shows strong improvement over the baseline's poor performance for the cross-lingual configurations. This demonstrates that BEPA's more robust alignment facilitates extractive question answering across language boundaries.

For the GLUE (General Language Understanding Evaluation) benchmark, we evaluate whether BEPA finetuning maintains performance on a range of English sentence-level understanding tasks. We finetune our pretrained model for three epochs on each GLUE task using the standard single linear layer and evaluation protocols \citep{wang2019glue}. GLUE comprises of classification and regression tasks, including natural language inference, paraphrase detection, sentiment analysis, and semantic similarity. Results for our models are found in \cref{tab:glue_results}.

\label{better transfer}

\FloatBarrier
\section{Conclusion}
\label{conclusion}
We release our new training framework for BERT-style models, BERT-JEPA (BEPA). It leverages a new training framework that combines classic BERT masked language modeling (MLM) with JEPA. When finetuning our BERT-style models under this new framework, we find the following:
(1) \textbf{BEPA finetuning drastically reorganizes the [CLS] embedding space to a semantic-first structure} (2) \textbf{BEPA finetuning shifts the PCA representation from low-rank to fuller-rank} (3) \textbf{this reorganization comes at little to no loss in English performance} and (4) \textbf{increases performance on multilingual tasks}.


\FloatBarrier

\FloatBarrier
\newpage
\bibliography{references}
\bibliographystyle{icml2022}

\newpage
\appendix
\onecolumn


\section{Appendix}
\label{appendix}

In this section, we share additional details and information pertaining to our experiments and methods. The appendix is laid out in the following way: Author contributions in \cref{contributions}, GenAI use in \cref{genai}, resources used in \cref{resources}, Our BEPA repository and code in \cref{repo}, the full BEPA framework in \cref{full bepa framework}, information on the collapsed BERT [CLS] embedding space in \cref{collapsed}, graphs pertaining to the training process \cref{training process}, results showing our new BERT embedding space \cref{new embedding space}, extensive results in \cref{more results}, and future work \cref{future work}. 

\subsection{Contributions}
\label{contributions}

In this section we outline the specific contributions of each author. These are in a randomized order that was decided by spinning a wheel with equal probabilities.

\textbf{Marcel's} main contributions to the paper include helping to come up with the BEPA framework and theoretical foundations, leading the writing of the paper, running evaluation on benchmarks, and creating plots. 

\textbf{Adam's} main contributions to the paper include helping to come up with the BEPA framework and theoretical foundations, leading evaluations, creating plots, and writing the paper.

\textbf{Taj's} main contributions to the paper include helping to come up with the BEPA framework and theoretical foundations, leading coding and model training, evaluations, and writing the paper.

\textbf{Kenneth's} main contributions to the paper include helping to come up with the BEPA framework and theoretical foundations, evaluations, and writing the paper.

\subsection{GenAI Use}
\label{genai}

GenAI was used for a couple of tasks during the development of this paper. It was leveraged as a tool for checking the grammar of our writing and correct it when needed. In addition, we used it streamline the code implementation portion, helping with debugging and the implementing our BEPA framework.

\subsection{Resources Used}
\label{resources}

In this section, we share different resources that were used throughout our paper. We begin by sharing the datasets used for finetuning and continue by sharing the datasets used for benchmarking our model.

\begin{table}[H]
  \label{dataset-table}
  \centering
\begin{minipage}{0.9\linewidth}
  \caption{Information on Datasets Used for training our different models across ablations. The number of language pairs represents the number of language combinations that translation samples existed for.}
  \begin{tabular}{|l|c|}
    \toprule
    \cmidrule(r){1-2}
    Name & Number of Language Pairs \\
    \midrule
    Opus-100 \citep{tiedemann-2012-parallel,zhang-etal-2020-improving}  & 114 \\
    WMT19 \citep{wmt19translate}  & 9  \\
    Flores \citep{nllb2022,flores2,flores3} & 200 \\
    Eng-Swa-Dataset \citep{Rogendo2025EngSwaDataset} & 2 \\
    \bottomrule
  \end{tabular}
  \end{minipage}
\end{table}

We found that Opus-100 did not have any English-Swahili pairs, so we used the Eng-Swa-Dataset as a bootstrapped method for allowing our model to learn relationships between those languages.

We wanted to make sure that we evaluated BEPA on a variety of downstream tasks. For this reason, we chose the following benchmarks to compare our finetuned models agains:
\begin{enumerate}
    \item  XNLI \citep{conneau2018xnli}
    \item  MLQA \citep{lewis2019mlqa}
    \item  GLUE \citep{agirre2007semantic, bar2006second, bentivogli2009fifth, dagan2006pascal, dolan2005automatically, giampiccolo2007third, levesque2011winograd, rajpurkar2016squad, socher2013recursive, wang2019glue, warstadt2018neural, williams2018broad} 
    
\end{enumerate}

\begin{table}[H]

  \label{dataset-table}
  \centering
  \begin{minipage}{0.82\linewidth}
  \caption{Benchmarks used to evaluate our models. These are the standard BERT evaluation benchmarks. \textbf{They span multiple different tasks, ranging from reasoning, analysis,  cross-lingual classification, and question and answering.}}
  \begin{tabular}{|l|c|}
    \toprule
    \cmidrule(r){1-2}
    Name & Task(s) \\
    \midrule
    XNLI  & Cross-lingual sentence classification \\    
    MLQA & Cross-lingual question answering (QA) \\    
    GLUE & Grammatical correctness, Sentiment analysis, Semantic similarity, Semantic reasoning \\ 
    \bottomrule
  \end{tabular}
  \end{minipage}
\end{table}

Due to GPU constraints, we did not run full model pretraining for this paper. However, for future work, we believe it is essential for a model to undergo full pretraining utilizing our new training framework. For finetuning, we utilized 1 NVIDIA RTX A5500 GPU, while training for 10 epochs. We leveraged a batch size of 16 and sample max length of 256 tokens to prevent running out of GPU memory. We applied a learning rate of $2x10^{-5}$ to an AdamW optimizer with a weight decay of 0.01 and 500 warmup steps. We used a $\lambda$ of 1 for out alginment loss and used the industry standard of masking probability of 15\% for the MLM task.

\FloatBarrier
\subsection{BEPA Repository}
\label{repo}

We release an anonymous version of our repository containing  the code for training BEPA. This includes the training, set up, configurations, and  training scripts. It can be found at the following url: \url{https://anonymous.4open.science/r/bert-jepa-translation-3EB8/README.md}

\FloatBarrier
\subsection{The Full BEPA Framework}
\label{full bepa framework}

Our novel BERT training method transforms the current embedding space for BERT models. It is made up of two main components:
\begin{enumerate}
    \item Classic MLM Training
    \item CLS-token JEPA Alignment
\end{enumerate}

\begin{figure}[]
  \centering
  \includegraphics[width=0.7\textwidth]{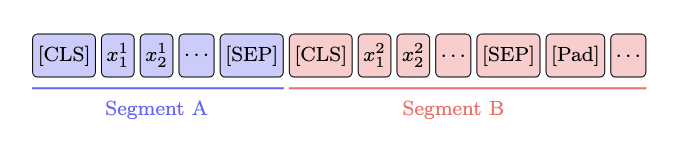}
  \caption{Input packing strategy for BEPA. Two sequences are concatenated using two \texttt{[CLS]} tokens and a \texttt{[SEP]} boundary. $x^{i}_1$ represents tokens corresponding to the first sequence while $x^{i}_2$ represents tokens corresponding to the second sequence. Segment A marks the first sentence and Segment B the second. \textbf{The entire sequence is encoded jointly.}}
  \label{tokenized input}
\end{figure}

We begin by taking the a base pretrained RoBERTa model (xlm-roberta-base) \citep{DBLP:journals/corr/abs-1911-02116}. Because we want to teach BERT to represent information independently of linguistic system, we choose to embed two sentences side by side, packing them into a single transformer input. We used the XLM-RoBERTa tokenizer and added special tokens to the samples such as [CLS], [SEP], and [PAD]. An example of the input can be found in figure \cref{tokenized input}. 

This leaves two possibilities for packaging: different samples in the same language or two samples in different languages packaged together. We took these possibilities and turned them into two separate training modes:
\begin{enumerate}
    \item Monolingual: Where the samples packaged together exclusively come from the same language.
    \item Bilingual: Where the samples packaged together can come from any language pairs (as long as they share a meaning).

\end{enumerate}

This setup then undergoes BERT's classic MLM task from the original BERT paper \citep{devlin2019bertpretrainingdeepbidirectional}. We do this to preserve the strong linguistic capabilities of BERT-style transformers. This step includes randomly masking a certain proportion of tokens, which is decided by a hyperparameter that was left at the proportion common to BERT: 15\%, predicting the masked tokens, and then taking MLM loss.

\begin{figure}[t]
  \centering
    \begin{subfigure}[b]{0.45\textwidth}
    \centering
    \includegraphics[width=\textwidth]{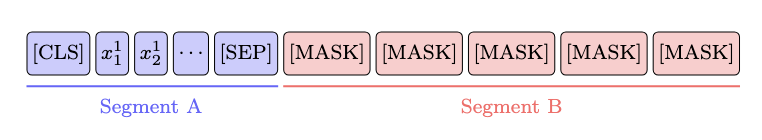}
    \label{fig:1-first}
  \end{subfigure}
  \hfill
  \begin{subfigure}[b]{0.49\textwidth}
    \centering
    \includegraphics[width=\textwidth]{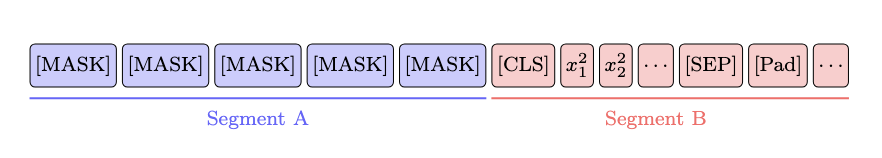}
    \label{fig:1-second}
  \end{subfigure}
  \hfill
  \vspace{-0.5cm}
  \caption{Masking strategy for CLS-token JEPA alignment stage of the BEPA framework. \textit{(Left)}: Masking out tokens corresponding to the second sentence to allow the for the model to build a [CLS] token representation of the first sequence. \textit{(Right)}: \textbf{Masking out tokens corresponding to the first sentence to allow for the model to build a [CLS] token representation of the second sequence.}
    }
  \label{BEPA masking}
\end{figure}

\begin{figure}[H]
  \centering
  \includegraphics[width=0.7\textwidth]{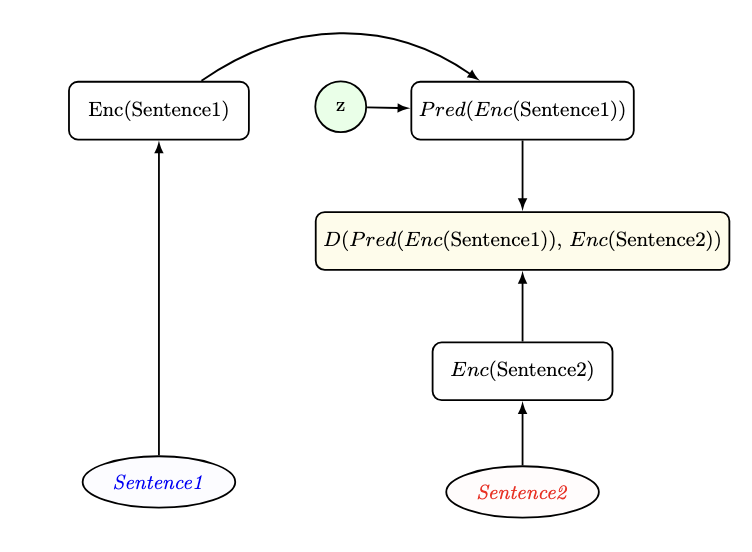}
  \caption{Architecture of the JEPA section of the BEPA framework. Practically, this is achieved through two forward passes of the BERT-style model with masking of the other sentence. The encoded output is the embedding of the [CLS] token corresponding to each sentence and the predictor is the identity matrix, leaving the latent representation the same. The final step calculates the distance between the two [CLS] latent representations and minimizes it.}
  \label{bepa architecture}
\end{figure}

We then move on to the novel part of the BEPA framework. Because BERT-style models treat the final-layer [CLS] token as the latent representation of the sequences, we take the [CLS] tokens and compute an alignment loss that encourages invariance across languages. To achieve this, we conducted two separate forward passes with each individual sentence masked out. The masking strategy can be found in \cref{BEPA masking}. The latent representation of each sentence's [CLS] token is then taken and aligned through an $L_{Alignment}$. We utilize an identity predictor; however, future work should test the impact of different predictors. A full diagram of the CLS-token alignment stage of BEPA can be found in \cref{bepa architecture}

The two steps, BERT-style MLM and CLS-token JEPA alignment result in a new training loss of:
$$L_{BEPA} = L_{MLM} + \lambda L_{Alignment} $$

in which $\lambda$ is a hyperparameter determining how much weight to put on $L_{Alignment}$, the loss used for the JEPA portion to align the [CLS] embeddings between the two samples with the same semantics. Because the [CLS] embeddings space of BERT-style models is inherently collapsed, we decided to use a loss that would help the model to separate negative pairs while bringing the representations of the similar pairs closer together. InfoNCE \citep{oord2019representationlearningcontrastivepredictive} and SigREG \citep{balestriero2025lejepaprovablescalableselfsupervised} were the most effective at fulfilling these requirements, but the framework is fully extensible and can be swapped out with whatever loss the user requires. Furthermore, experiments were conducted using an identity function predictor but in future work, we believe it to be important to test the impact that different predictors have on BEPA.

\FloatBarrier

\subsection{A Collapsed CLS Token}
\label{collapsed}

In this section, we look at how the [CLS] tokens for BERT models are collapsed, resulting in a latent space that loses a lot of valuable information. If the [CLS] token is used for most downstream tasks for BERT embeddings, then it is imperative that the latent space produced by the [CLS] tokens is rich and not collapsed. This collapse is perfectly exemplified by the cosine similarity and euclidean distance between different pairs of points when embedded by BERT-style models. Resulting in a situation where all sentences, even dissimilar pairs, have high cosine similarity. This can be seen in \cref{collapsed-cls-full-roberta,collapsed-cls-full-xlm}.





\begin{table}[H]
\centering
\begin{minipage}{0.86\linewidth}
\centering
\caption{Cosine similarity between embeddings of sentences across and within languages for \textit{RoBERTa} for wmt19 dataset. 
\textbf{There is a high cosine similarity between all sentence pairs, even unrelated and cross-language pairs. Showing a collapsed [CLS] token embedding space as they are all highly similar.}}
\label{collapsed-cls-full-roberta}

\begin{tabular}{|l|ccc|}
\toprule
\textbf{Language Pair} &
\textbf{Same Language (unrelated)} &
\textbf{Diff Language (related)} &
\textbf{Diff Languages (unrelated)} \\
\midrule
ar--en & 0.9968 & 0.9918 & 0.9935 \\
bg--en & 0.9965 & 0.9941 & 0.9937 \\ 
de--en & 0.9959 & 0.9965 & 0.9926 \\ 
el--en & 0.9965 & 0.9923 & 0.9936 \\ 
en--es & 0.9961 & 0.9973 & 0.9928 \\ 
en--fr & 0.9957 & 0.9964 & 0.9919 \\ 
en--hi & 0.9961 & 0.9908 & 0.9909 \\ 
en--ru & 0.9965 & 0.9929 & 0.9933 \\ 
en--th & 0.9976 & 0.9904 & 0.9933 \\ 
en--tr & 0.9961 & 0.9952 & 0.9937 \\ 
en--ur & 0.9968 & 0.9887 & 0.9917 \\ 
en--vi & 0.9963 & 0.9939 & 0.9931 \\ 
en--zh & 0.9959 & 0.9919 & 0.9932 \\ 
\bottomrule
\end{tabular}

\end{minipage}
\end{table}


\begin{table}[H]
\centering
\begin{minipage}{0.86\linewidth} 
\centering

\caption{Cosine similarity between embeddings of sentences across and within languages for \textit{XLM-RoBERTa} for wmt19 dataset. 
\textbf{There is a high cosine similarity between all sentence pairs, even unrelated and cross-language pairs. Showing a collapsed [CLS] token embedding space as they are all highly similar.}}
\label{collapsed-cls-full-xlm}


\begin{tabular}{|l|ccc|}
\toprule
\textbf{Language Pair} &
\textbf{Same Language (unrelated)} &
\textbf{Diff Language (related)} &
\textbf{Diff Languages (unrelated)} \\
\midrule
ar--en & 0.9923 & 0.9926 & 0.9979 \\
bg--en & 0.9957 & 0.9945 & 0.9845 \\
de--en & 0.9923 & 0.9968 & 0.9923 \\
el--en & 0.9929 & 0.9959 & 0.9925 \\
en--es & 0.9918 & 0.9970 & 0.9939 \\
en--fr & 0.9877 & 0.9949 & 0.9907 \\
en--hi & 0.9970 & 0.9981 & 0.9903 \\
en--ru & 0.9910 & 0.9975 & 0.9916 \\
en--th & 0.9975 & 0.9956 & 0.9927 \\
en--tr & 0.9924 & 0.9962 &  0.9891 \\
en--ur & 0.9984 & 0.9977 & 0.9928 \\
en--vi & 0.9939 & 0.9961 & 0.9918 \\
en--zh & 0.9967 & 0.9981 & 0.9900 \\
\bottomrule
\end{tabular}

\end{minipage}
\end{table}

\FloatBarrier
\subsection{The Training Process}
\label{training process}
\begin{figure}[H]
    \centering
    
    \begin{subfigure}[b]{0.33\textwidth}
        \centering
        \includegraphics[width=\textwidth]{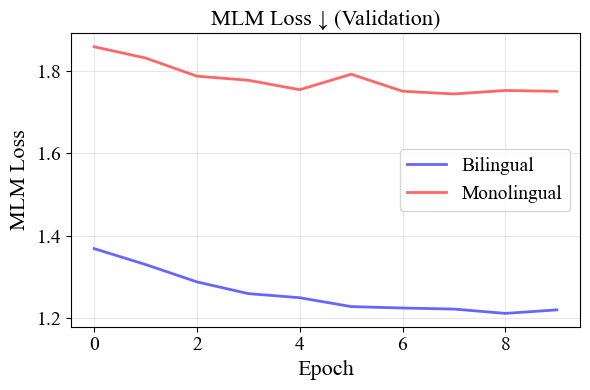}
        \label{fig:mlm_loss}
    \end{subfigure}
    \hfill
    \begin{subfigure}[b]{0.33\textwidth}
        \centering
        \includegraphics[width=\textwidth]{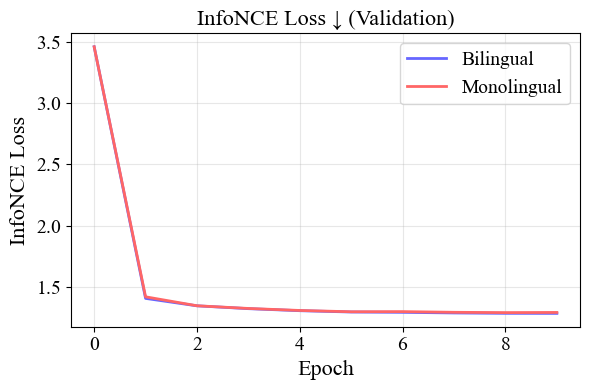}
        \label{fig:infonce_loss}
    \end{subfigure}    
    \hfill
    \begin{subfigure}[b]{0.33\textwidth}
        \centering
        \includegraphics[width=\textwidth]{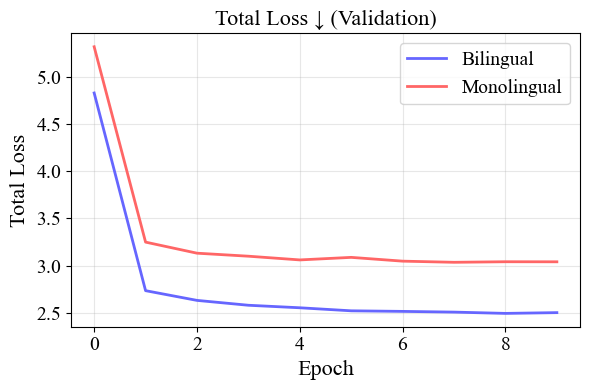}
        \label{fig:total_loss}
    \end{subfigure}
    \caption{Validation losses across training epochs. \textit{(Left)} Masked language modeling validation loss. The bilingual model has consistently lower loss, as it is given the sentence in two languages and may have the direct translation for the masked token available. \textit{(Middle)} InfoNCE validation loss. This is taken across each batch, using the translation as a positive and all other sentences as negatives. This is not dependent on the MLM task. \textit{(Right)} Total validation loss across training epoch.}
    \label{fig:losses}
\end{figure}

\begin{figure}[H]
  \centering
  \includegraphics[width=0.48\textwidth]{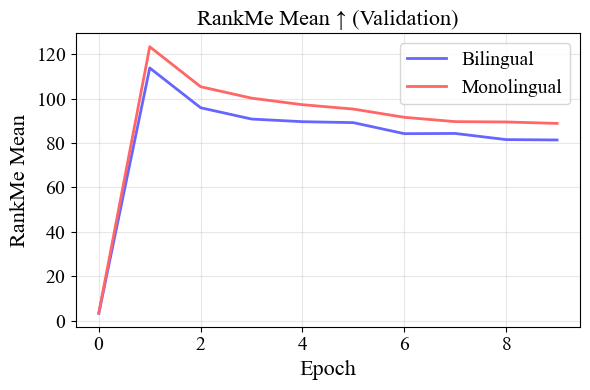}
  \caption{RankMe \citep{garrido2023rankmeassessingdownstreamperformance} average of the [CLS] token for both languages in the bilingual task or the [CLS] token alone in the monolingual task taken on the validation set across epochs. RankMe gives the effective rank of the embeddings. The model starts with a collapsed space, spiking briefly, then settling. Monolingual training exhibits a higher RankMe across training than bilingual training.}
  \label{fig:rankme}
\end{figure}

In this section, we examine the training processes, focusing on the loss (Figure \ref{fig:losses}) and RankMe \citep{garrido2023rankmeassessingdownstreamperformance} (Figure \ref{fig:rankme}) values across epochs.

\FloatBarrier
\subsection{A Richer Embedding Space}
\label{new embedding space}
In this section, we cover the results related to a richer [CLS] embedding space post BEPA finetuning. \cref{uncollapsed-cls-full-mono,uncollapsed-cls-full-bilingual} show the resulting cosine similarities between related and unrelated sentences, a big improvement from the collapsed [CLS] tokens from \cref{collapsed}.

\begin{figure}[H]
  \centering
  \includegraphics[width=0.7\textwidth]{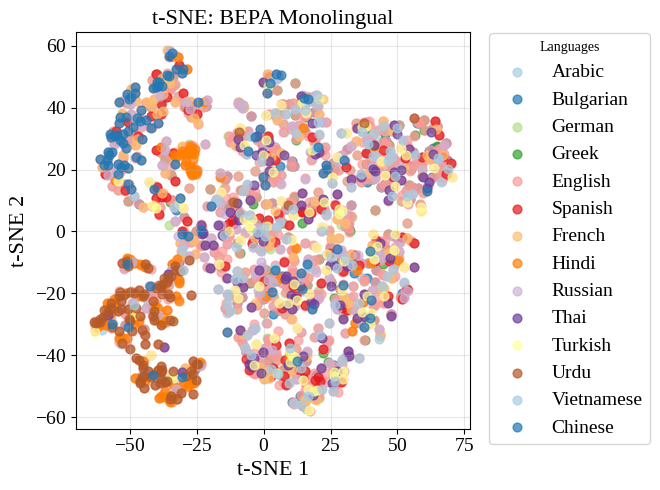}
  \caption{t-SNE plot for BEPA Monolingual showing evenly distributed CLS token embeddings across the space. \textbf{BEPA creates a shared and aligned space between languages}. }
  \label{tsne_plot_mono}
\end{figure}

\begin{figure*}[]
  \centering
  \begin{subfigure}{0.48\textwidth}
    \centering
    \includegraphics[width=\linewidth]{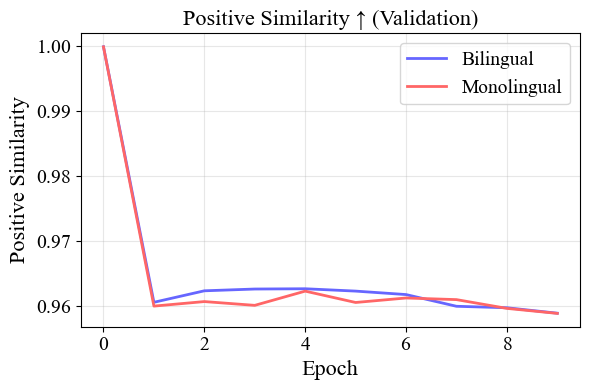}
    \label{fig:1-pos}
  \end{subfigure}
  \hfill
  \begin{subfigure}{0.48\textwidth}
    \centering
    \includegraphics[width=\linewidth]{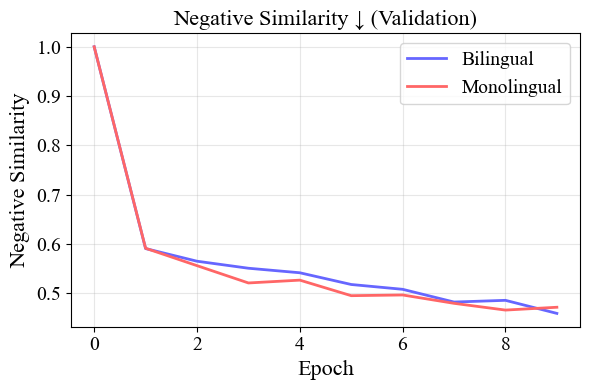}
    \label{fig:1-neg}
  \end{subfigure}
\vspace{-20pt}
  \caption{Mean Cosine similarities as a relation of epoch for positive and negative samples. \textit{(Left)}: Mean Cosine Similarity of related (similar semantically) validation-set samples remains high post BEPA finetuning. \textit{(Right)}: Mean Cosine Similarity of unrelated (dissimilar semantically) validation-set samples decreases (↓) post BEPA finetuning.
  \textbf{Overall, BEPA finetuning reorganizes the [CLS] embeddings space and fights collapse. Similar pairs are close together and dissimilar pairs are further apart.} For further examples, please refer to \cref{uncollapsed-cls-full-bilingual,uncollapsed-cls-full-mono}.}
  \label{fig:pos-neg}
\end{figure*}

\begin{table}[]
\centering
\begin{minipage}{0.86\linewidth}
\centering
\caption{Cosine similarity between embeddings of sentences across and within languages for \textit{BEPA-Monolingual} for wmt19 dataset. 
There is a high (↑) cosine similarity between similar sentence pairs, and low between unrelated pairs. Showing an un-collapsed [CLS] token embedding space. \textbf{BEPA training decreases (↓) cosine similarity between unrelated pairs, leading to a richer embedding space.}}
\label{uncollapsed-cls-full-mono}

\begin{tabular}{|l|ccc|}
\toprule
\textbf{Language Pair} &
\textbf{Same Language (unrelated)} &
\textbf{Diff Language (related)} &
\textbf{Diff Languages (unrelated)} \\
\midrule
ar--en & 0.5683 & 0.9719 & 0.5083 \\
bg--en & 0.5986 & 0.9641 & 0.4804 \\
de--en & 0.4940 & 0.9739 & 0.4445 \\
el--en & 0.4709 & 0.9751 & 0.4687 \\
en--es & 0.4877 & 0.9885 & 0.4506 \\
en--fr & 0.3797 & 0.9815 & 0.4163 \\
en--hi & 0.5388 & 0.9875 & 0.4451 \\ 
en--ru & 0.4113 & 0.9716 & 0.4693 \\
en--th & 0.6198 & 0.9568 & 0.4954 \\ 
en--tr & 0.5656 & 0.9751 & 0.3966 \\ 
en--ur & 0.6758 & 0.9878 & 0.3952 \\ 
en--vi & 0.5783 & 0.9695 & 0.5595 \\ 
en--zh & 0.5087 & 0.9824 & 0.3713 \\

\bottomrule
\end{tabular}
\end{minipage}
\end{table}

\begin{table}[]
\centering
\begin{minipage}{0.86\linewidth}
\centering
\caption{Cosine similarity between embeddings of sentences across and within languages for \textit{BEPA-Bilingual} for wmt19 dataset. 
There is a high (↑) cosine similarity between similar sentence pairs, and low between unrelated pairs. Showing an un-collapsed [CLS] token embedding space. \textbf{BEPA training decreases (↓) cosine similarity between unrelated pairs, leading to a richer embedding space.}}
\label{uncollapsed-cls-full-bilingual}
\begin{tabular}{|l|ccc|}
\toprule
\textbf{Language Pair} &
\textbf{Same Language (unrelated)} &
\textbf{Diff Language (related)} &
\textbf{Diff Languages (unrelated)} \\
\midrule
ar--en & 0.5351 & 0.9719 & 0.5035 \\
bg--en & 0.5786 & 0.9665 & 0.5207 \\ 
de--en & 0.4706 & 0.9757 & 0.4718 \\ 
el--en & 0.4911 & 0.9755 & 0.4336 \\ 
en--es & 0.4640 & 0.9897 & 0.4158 \\ 
en--fr & 0.4600 & 0.9829 & 0.4147 \\ 
en--hi & 0.5080 & 0.9862 & 0.4290 \\ 
en--ru & 0.4522 & 0.9712 & 0.3601 \\ 
en--th & 0.6369 & 0.9553 & 0.5377 \\ 
en--tr & 0.5629 & 0.9759 & 0.4996 \\ 
en--ur & 0.6757 & 0.9876 & 0.3838 \\ 
en--vi & 0.6027 & 0.9695 & 0.5331 \\ 
en--zh & 0.4345 & 0.9825 & 0.3117 \\
\bottomrule
\end{tabular}
\end{minipage}
\end{table}



\FloatBarrier
\subsection{Ablation Benchmark Results}
\label{more results}

In this section, we include additional results from different ablations that we ran. These include results for the following benchmarks: GLUE (\cref{tab:glue_results}), XNLI (\cref{tab:xnli_results_app}), MLQA (\cref{tab:mlqa_results}). We investigated how data quality and dataset size influence cross-lingual transfer performance. We fine-tuned XLM-RoBERTa-Base on three different training configurations: (1) Opus models used 10,000 translation pairs from the Opus-100 dataset and Swahili-English translations dataset \citep{tiedemann-2012-parallel, tiedemann2020tatoebatranslationchallenge, Rogendo2025EngSwaDataset}, (2) Flores models used approximately 2,000 pairs from the higher-quality Flores dataset with 5× more training epochs to maintain comparable gradient update steps \citep{nllb2022, flores2, flores3}, and (3) Small models were trained on 2,000 examples from Opus-100 and with 5x more training epochs to maintain comparable gradient update steps All models were evaluated on the XNLI benchmark to assess zero-shot cross-lingual transfer capabilities \citep{tiedemann-2012-parallel, tiedemann2020tatoebatranslationchallenge, Rogendo2025EngSwaDataset}.

\begin{table}[H]
\centering
\begin{minipage}{0.76\linewidth} 
\caption{Average results across 5 seeds on the GLUE benchmark. We report accuracy for all classification tasks and Pearson correlation for STS-B. Results with $\dagger$ are from the original XLM-R paper \citep{DBLP:journals/corr/abs-1911-02116}. We also report our results from GLUE with XLM-RoBERTa on our own hardware and training/testing configuration. \textbf{BEPA finetuning leads to little to no loss in performance on English tasks}.}
\small
\begin{tabular}{|l|cccccc|c|}
\toprule
\textbf{Model} & \textbf{MNLI-m/mm} & \textbf{QNLI} & \textbf{QQP} & \textbf{SST} & \textbf{MRPC} & \textbf{STS-B} & \textbf{Avg} \\
\midrule
XLM-R$^\dagger$ & 88.9 / 89.0 & 93.8 & 92.3 & 95.0 & 89.5 & 91.2 & 91.8 \\
\midrule
XLM-RoBERTa (Ours) & 84.64 / 84.59 & 90.38 & 90.69 & 92.55 & 86.32 & 87.36 & 88.07 \\
JEPA Monolingual & 84.31 / 84.47 & 90.57 & 90.60 & 92.87 & 86.47 & 87.56 & 88.12 \\
JEPA Bilingual & 84.97 / 85.04 & 90.82 & 90.83 & 92.20 & 88.24 & 88.39 & 88.64 \\
\bottomrule
\end{tabular}
\label{tab:glue_results}
\end{minipage}
\end{table}

\begin{table}[htbp]
\centering
\caption{Average accuracy (decimal) across 5 seeds on XNLI Language Transfer benchmark, displayed by language and overall. Displaying results for all our models, ablations tested, and \textit{XLM-RoBERTa base}. Results are gathered by finetuning on English and then doing zero-shot evaluation on each language. \textbf{\textit{BEPA Bilingual} consistently outperforms (↑) \textit{XLM-RoBERTa} across all languages.} For additional results and ablations refer to \cref{tab:xnli_results,more results}.}
\label{tab:xnli_results_app}
\resizebox{\textwidth}{!}{%
\begin{tabular}{|l|ccccccccccccccc|c|}
\toprule
\textbf{Model} & \textbf{ar} & \textbf{bg} & \textbf{de} & \textbf{el} & \textbf{en} & \textbf{es} & \textbf{fr} & \textbf{hi} & \textbf{ru} & \textbf{sw} & \textbf{th} & \textbf{tr} & \textbf{ur} & \textbf{vi} & \textbf{zh} & \textbf{Avg} \\
\midrule
XLM-RoBERTa Base & 0.705 & 0.753 & 0.753 & 0.744 & 0.831 & 0.783 & 0.767 & 0.687 & 0.740 & 0.610 & 0.714 & 0.706 & 0.656 & 0.735 & 0.741 & 0.728 \\
\midrule
Bilingual - Opus & 0.717 & 0.765 & 0.771 & 0.763 & 0.842 & 0.794 & 0.780 & 0.693 & 0.750 & 0.673 & 0.726 & 0.725 & 0.657 & 0.749 & 0.751 & 0.744 \\
Bilingual - Flores & 0.718 & 0.764 & 0.764 & 0.757 & 0.841 & 0.788 & 0.778 & 0.693 & 0.746 & 0.656 & 0.718 & 0.719 & 0.656 & 0.742 & 0.745 & 0.739 \\
Bilingual - Small & 0.710 & 0.760 & 0.769 & 0.757 & 0.839 & 0.792 & 0.776 & 0.691 & 0.750 & 0.664 & 0.719 & 0.729 & 0.658 & 0.746 & 0.747 & 0.741 \\
\midrule
Monolingual - Opus & 0.707 & 0.755 & 0.755 & 0.745 & 0.829 & 0.782 & 0.771 & 0.681 & 0.738 & 0.662 & 0.714 & 0.718 & 0.654 & 0.735 & 0.735 & 0.732 \\
Monolingual - Flores & 0.701 & 0.751 & 0.757 & 0.743 & 0.829 & 0.782 & 0.768 & 0.687 & 0.740 & 0.652 & 0.710 & 0.709 & 0.655 & 0.735 & 0.736 & 0.730 \\
Monolingual - Small & 0.700 & 0.750 & 0.751 & 0.741 & 0.830 & 0.779 & 0.768 & 0.685 & 0.733 & 0.660 & 0.718 & 0.721 & 0.653 & 0.737 & 0.731 & 0.730 \\
\bottomrule
\end{tabular}%
}
\end{table}

\begin{table}[H]
\caption{Results on MLQA question answering. We report F1/EM scores for zero-shot classification where models are fine-tuned on English SQuAD and evaluated on 7 languages. Results with $\dagger$ are from the original MLQA paper \citep{lewis2019mlqa}.}
\centering
\small
\resizebox{\textwidth}{!}{
\begin{tabular}{|l|ccccccc|c|}
\toprule
\textbf{Model} & \textbf{en} & \textbf{es} & \textbf{de} & \textbf{ar} & \textbf{hi} & \textbf{vi} & \textbf{zh} & \textbf{Avg} \\
\midrule
BERT-Large$^\dagger$ & 80.2/67.4 & - & - & - & - & - & - & - \\
mBERT$^\dagger$ & 77.7/65.2 & 64.3/46.6 & 57.9/44.3 & 45.7/29.8 & 43.8/29.7 & 57.1/38.6 & 57.5/37.3 & 57.7/41.6 \\
XLM-15$^\dagger$ & 74.9/62.4 & 68.0/49.8 & 62.2/47.6 & 54.8/36.3 & 48.8/27.3 & 61.4/41.8 & 61.1/39.6 & 61.6/43.5 \\
XLM-R$_{\text{Base}}$ & 77.1/64.6 & 67.4/49.6 & 60.9/46.7 & 54.9/36.6 & 59.4/42.9 & 64.5/44.7 & 61.8/39.3 & 63.7/46.3 \\
XLM-R & 80.6/67.8 & 74.1/56.0 & 68.5/53.6 & 63.1/43.5 & 69.2/51.6 & 71.3/50.9 & 68.0/45.4 & 70.7/52.7 \\
\midrule
XLM-RoBERTa (Ours) & 80.9/68.2 & 66.8/45.8 & 62.4/46.8 & 55.0/35.9 & 61.7/44.0 & 67.4/47.0 & 40.3/39.6 & 62.1/46.8 \\
BEPA Monolingual (Ours) & 81.0/68.0 & 67.8/46.1 & 62.4/46.4 & 55.5/36.6 & 64.0/46.0 & 68.6/47.6 & 41.3/40.6 & 62.9/47.3 \\
BEPA Monolingual (Ours) & 81.2/68.2 & 67.8/46.4 & 63.3/47.2 & 55.9/37.0 & 63.2/45.4 & 68.5/47.8 & 42.3/41.6 & 63.2/47.7 \\
\bottomrule
\end{tabular}}
\label{tab:mlqa_results}
\end{table}

\subsection{Future Work}
\label{future work}

This is only the beginning for BEPA. Several experiments remain to be done. Here, we highlight the future work and experiments that we are interested in running: (1) We want to see the impact of a different alignment loss, such as SigREG during finetuning. (2) Testing different weights ($\lambda$) on the alignment loss (3) Extending our framework to  an autoregressive model, as autoregression implicitly prevents collapse (4) Experimenting with different predictors for the JEPA predictor such as a predictor network, and leveraging the use of latent variables.

\end{document}